\documentclass[10pt,twocolumn,letterpaper]{article}

\usepackage{iccv}
\usepackage{times}
\usepackage{epsfig}
\usepackage{graphicx}
\usepackage{amsmath}
\usepackage{amssymb}

\usepackage[breaklinks=true,bookmarks=false]{hyperref}

\iccvfinalcopy 

\def\iccvPaperID{****} 

\begin{document}

\title{The role of RGB-D benchmark datasets: an overview}

\author{Kai Berger\\
Oxford e-Research Centre\\
7 Keble Road, Oxford\\
{\tt\small kai.berger@oerc.ox.ac.uk}
}

\maketitle

\begin{abstract}
The advent of the Microsoft Kinect three years ago stimulated not only the computer vision community
for new algorithms and setups to address well-known problems in the community but also sparked the launch
of several new benchmark datasets to which future algorithms can be compared to.
   This review of the literature and industry developments
concludes that the current RGB-D benchmark datasets can be useful to determine
the accuracy of a variety of applications of a single or multiple RGB-D sensors.
\end{abstract}
\begin{figure*}
\centering
\includegraphics[height=.23\linewidth]{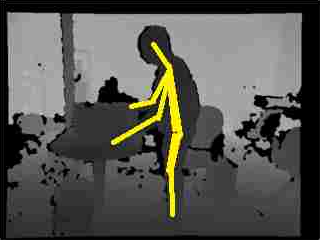}
\includegraphics[height=.23\linewidth]{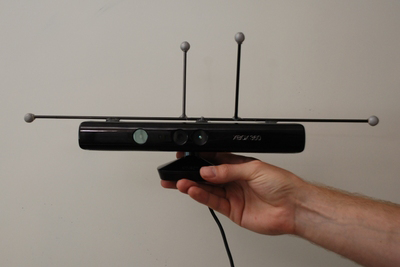}
\includegraphics[height=.23\linewidth]{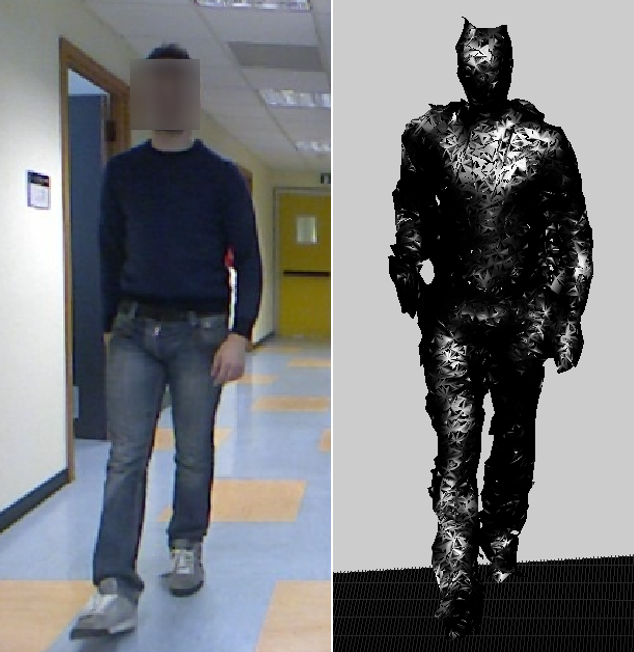}
\includegraphics[height=.23\linewidth]{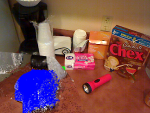}
\includegraphics[height=.23\linewidth]{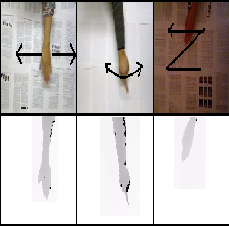}
\includegraphics[height=.23\linewidth]{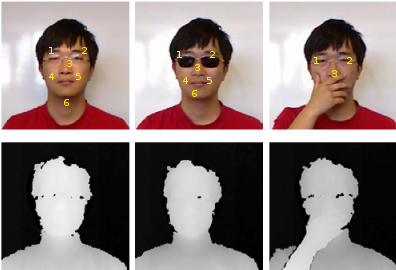}
\caption{\label{overviewfig} A collage of the variety of benchmark datasets that are currently publicly available. Top left: depth images with annotated motion (reproduced from \cite{sung2011human}), top middle: 
external tracking of Kinect pose with markers (reproduced from \cite{sturm2011towards}), top right: tight mesh and skeleton alongside rgb-data (reproduced from Barbosa \etal \cite{barbosa2012re}).
Bottom left: depth images with objects annotated (reproduced from \cite{lai2011large}), bottom middle: depth data with annotated hand movements (reproduced from \cite{liu2013learning}), bottom right: face capturings in RGB-D stream anotated (reproduced from Huynh \etal \cite{huynh2013efficient})
.}
\end{figure*}
\section{Introduction}
The commercial success of the Microsoft Kinect~\cite{pressRelease} in November 2010 sparked a multitude of significant research papers in the computer vision community. 
The Microsoft Kinect was originally designed as a motion sensing input device for the gaming console Microsoft XBOX 360 by tracking the player's motions. 
As structured light sensor, the Kinect emits a defined light spot pattern, which was first patented by PrimeSense.
Users can access the Kinect data streams via USB 2.0 with the help of OpenKinect's \textit{libfreenect}.
The main advantage of Kinect capturing setups over conventional time-of-flight (ToF) setups is that the cost is only a fraction of the usual ToF-setups, which makes experimenting with one or many Kinects very convenient.

The projected spot pattern, used for computing the depth maps, is generated as follows: an infrared laser, projects a defined pattern at $850nm$ onto all surfaces of the scene facing the sensor in the frustum.
The diffuse reflection of the pattern in the scene is captured by a camera, which has its infrared filter removed.
An onboard circuit computes the disparity for each $9 \times 9$ subpattern by computing the distance to their default positions for an image of a default scene (this is likely to be a wall parallel to the sensor at a defined distnce of $3$m). The disparity values are mapped to distance values in meters. This technique has been introduced by PrimeSense.

 The project pattern is a texture of $211 \times 165$ spot positions. 3861 spot positions are brighter, the rest is assumed dark. That pattern is replicated in a $3\times3$ pattern to broaden the field of view.
 The central spot of the pattern appears brighter than all the other spots and no two bright spots are adjacent.
The pattern looks quasi-random but in fact is the same for all cameras. Thus, one device can compute the depth map from the emitted pattern of another adjacent device, if its own laser is obstructed.
 It is assumed that the depth computation follows a block search approach, i.e. the integrated circuit looks the corresponding position in the neighbourhood of the original subpattern position and computs depth values from the distance to the original position.
The visual tact rate can be computed from a central distinguishable subpattern, where a horizontal line alternates bright and dark
spots. 
As several brighter spots are visible over the complete pattern, it can be assumed, that distortions might be calculated from their location information in the received image.

Located within the Kinect device, there is one RGB camera, that operates at 30Hz with a resolution of $640
\times 480$ pixels or 15Hz with a resolution of $1280
\times 1024$ pixels, and one IR camera, that operates at 30Hz with a resolution of $640
\times 480$ pixels or 10Hz with a resolution of $1280 \times 1024$ pixels.
In the IR view it can be noticed, that visible light is captured with the chip at a small intensity range as well, and that a Bayer pattern shows when zooming in.
It is assumed, that a diffractive element splits up the laser light before it traverses a mechanical grid with occluding material applied to the spots that are designed to be dark in the projection.
For completeness reasons it should be noted, that the Kinect also bears a microphone array and an accelerometer. The Kinect can be tilted on command via USB.
Most capturing environments presented in this report are indoors, such that daylight with intensity in the IR spectrum does not oversaturate the recorded image.

Together with Asus Xtion and PrimeSense depth acquisition has become significantly easier. Thanks to accurate depth data, currently published papers could present a broad range of RGB-D setups addressing well-known problems in computer vision in which the Microsoft Kinect ranging from SLAM \cite{henry2010rgb,hu2012rgbdslam,lieberknecht2011rgb,lee2012rgbd,smisek20113d,herrera2011accurate} over 3d reconstruction \cite{alexiadis2012reconstruction,rafibakhsh2012analysis,sumarfeasability,pancham2012mapping,ahmedsystem} over realtime face \cite{leyvand2011kinect} and hand \cite{oikonomidis2011efficient} tracking to motion capturing and gait analysis \cite{santhanam2011th,wilson2010combining,fuhrmannmulti,berger2011markerless,asteriadis2013estimating}.
The course of the research over the past years also required some datasets captured with the Microsoft Kinect or a similar RGB-D sensor to be made publicly available for comparison.

In the following sections we provide an overview over the significant benchmarks that are currently publicly available for comparison, Fig.~\ref{overviewfig} . A tabular overview about the corresponding publications for each dataset can found in Table~\ref{overviewtable}.
\subsection{Method Of Comparison}
\begin{figure*}
\centering
\includegraphics[width=.45\linewidth]{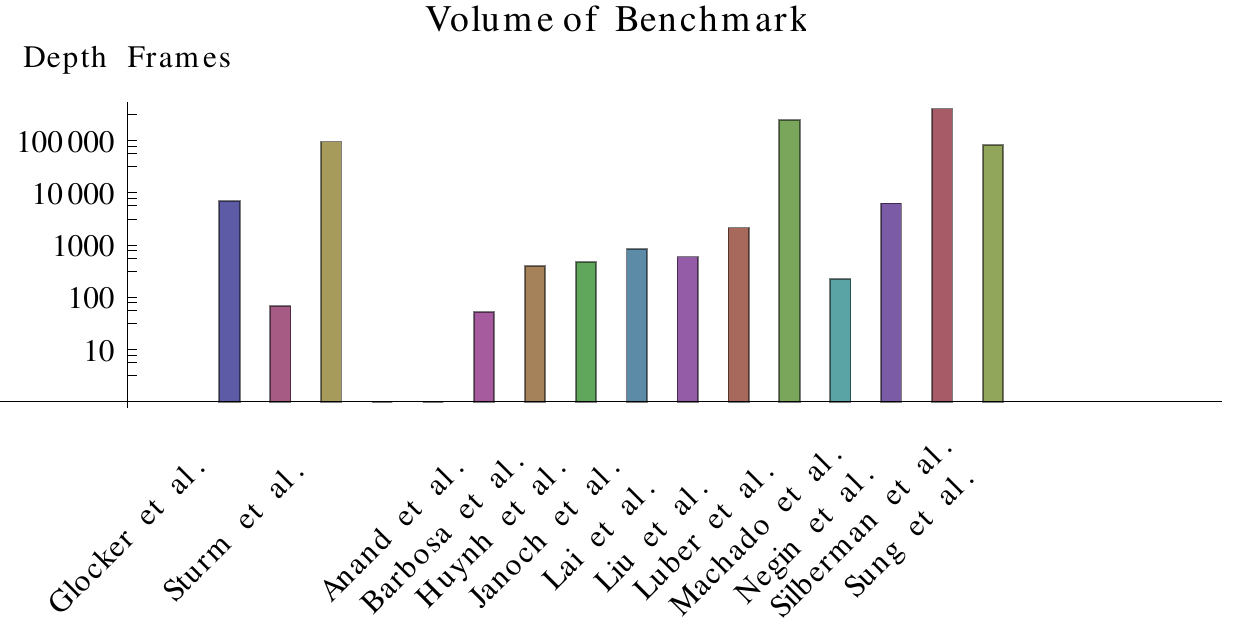}
\includegraphics[width=.45\linewidth]{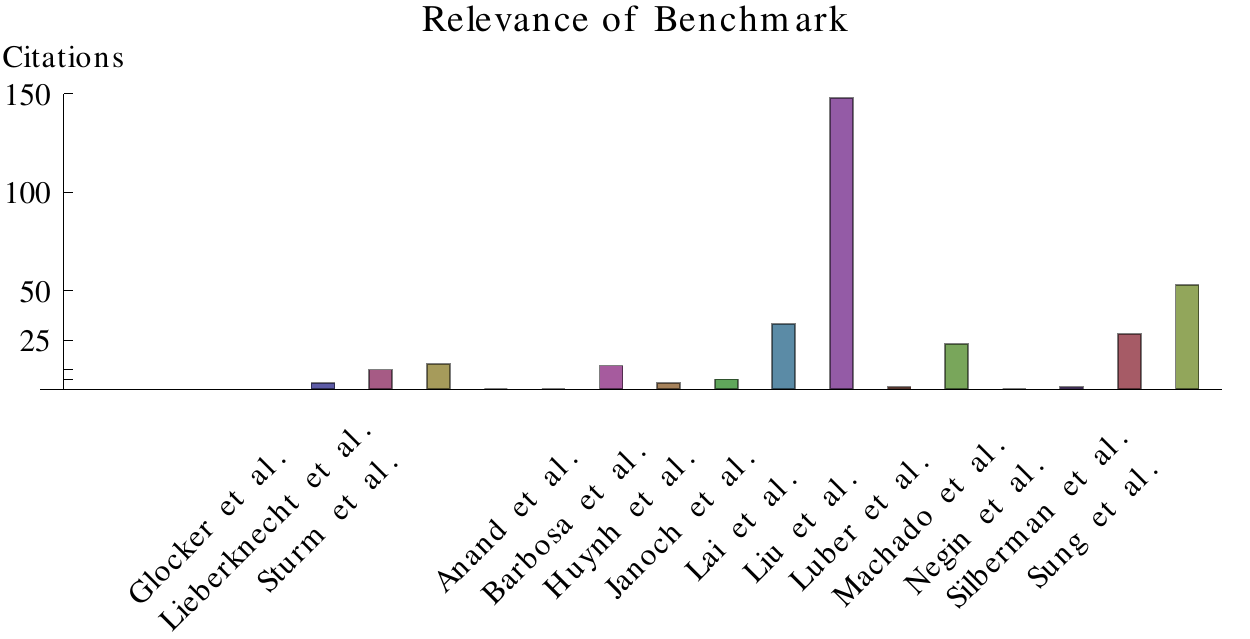}
\caption{\label{relevance} Left: This semi-logarithmic bar chart depicts the size of each published dataset in terms of absolute depth-images. The dataset presented by Silberman \etal \cite{Silberman:ECCV12} bears the most input images. Right: This chart depicts the impact of each published dataset in the community. It is sorted alphabetically for each research field. The work by Lai \etal \cite{lai2011large} has been considered most in the community.}
\end{figure*}

This overview paper does not provide new research findings but it attempts to provide an overview over the diverse set of benchmarks that are publicly available for comparison of RGB-D based algorithms
The findings are summarized in an overview table, Table~\ref{overviewtable} and compared for main distinguishable criteria. The table is sorted alphabetically for each research field, i.e. \emph{SLAM}, Sect.~\ref{slam} and \emph{Object Recognition}, Sect.~\ref{object}. We evaluated if the accelerometer of the Kinect was used (third column), if the data were annotated and which type of ground truth has been made available (fourth column). Finally we provided the link to the datasets (fifth columns). We tested the accessability in the middle of August. Some datasets may require login data, which however can be acquired by contacting the corresponding authors (intstructions were published on the corresponding website in that case). In Sect.~\ref{shortcomings} we provide a critical view onto the diversity of the publicly available datasets and phrase suggestions for extending the state of the art in benchmarks. Statistics about the volume and impact of each dataset is provided in Fig.~\ref{relevance}.

\section{Annotation for Ground Truth retrieval}
\label{annotation}
Most datasets exceed a feasible size to be handled by a single user for annotation. Hence, with the increasing popoluratiy of internet freelance websites, most publications presented in this report have relied on Mechanical Turk, e.g. \cite{janoch2013category}, for robust annotation of the datasets. Some rely on additional sensors to provide the ground truth, e.g. for the camera pose at a given frame~\cite{sturm2011towards,sturm2012benchmark}. A sophisticated approach transforms the labeling in another space: instead of letting the user annotate in image space, the static scene captured with a moving Kinect is reconstructed in 3d and annotated in a 3d graphics tool once, e.g.~\cite{lai2011large}. The annotated point clouds are then simply reprojected into the input stream using the camera pose for the Kinect sensor at each frame. 
\section{SLAM}
\label{slam}
Highly accurate depth data are necessary for 3D reconstruction and simultaneous reconstruction and simultaneous localization and mapping (SLAM) applications, although the requirements for mapping or localization can differ within the applicational context.
It can be seen, that accuracy and the running time/framerates trade each other off. The Kinect is the first device that provides fast data acquisition at acceptable accuracy.
In their work Sturm \etal \cite{sturm2011towards,sturm2012benchmark} release a 50 GB dataset conisisting of 39 RGB-D sequences captured with the Microsoft Kinect including the recorded accelerometer data with the intention to test SLAM algorithms on the input data. The authors provide ground truth via external per frame pose estimation of the Kinect within a global reference framework, which has been computed from the capturing of markers that have been attached to the Kinect beforehand. They used a MotionAnalysis capturing system at 100 Hz. Lieberknecht \etal \cite{lieberknechtIsmar11} create also a benchmark for localisation and provide video data, from which the RGB and depth data can be extracted. However, they do not provide a dataset that contains annotations or additional data, e.g. accelerometric data. Glocker \etal \cite{Glocker2013} prvoide a dataset captured with a moving camera and use KinectFusion to generate the 3d scene and the camera path as ground truth for the benchmark. They provide seven different scenes including RGB, depth and pose data in a txt-file.
\section{Object Recognition}
\label{object}
Based on the Kinect's realtime output of accurate depth maps, it became possible to reconstruct 3D objects with the Kinect, e.g. by moving the sensor around the acquired object.
For example, Tam and his colleagues~\cite{tam2012registration} register point clouds captured with the Kinect to each other.
Lai \etal \cite{lai2011large} present an annotated dataset containing visual and depth images of 300 physically distinct objects ranging from fruits to tools. Their dataset was captured with the Primesense prototype and a Firewire RGB-camera from Pointgrey. Their approach to labeling the objects in the input sequences is somewhat innovative: they reconstruct the 3d scene from the moving RGB-D sensor setup while keeping track of its position over time. The objects of interest are then labeled once in the 3d scene by hand and then backprojected into the input streams.
Liu \etal \cite{liu2013learning} present a dataset for gesture recognition where 2160 hand gesture sequences of 6 persons are captured with the Microsoft Kinect. The annotated dataset differentiates 10 hand gestures: circle (clockwise), triangle (anti-clockwise), up-down, right-left, wave, ”Z”, cross, comehere, turnaround. As the Microsoft Kinect remains fixed during acquisition there is no additional accelerometric data in the dataset. Negin \etal \cite{negin2013decision} provide a dataset of human body movements represented by 3D positions of skeletal joints. As the Kinect sensor remained fixed, no accelerometric data is available, but the authors provide the complete trackign resutls gained from applying the Microsoft Kinect SDK to the RGB-D data as the ground truth for their benchmark. In the dataset 15 people conduct 10 different exercises. Barbosa \etal \cite{barbosa2012re} capture 79 persons first for a distinctive signature, e.g. in a defined pose, and then in regular motion, e.g. walking across a floor. They provide both skeleton fits and .ply meshes alongside the RGB-D data. The goal of their dataset is to reindentify different humans captured with the Kinect. The humans may change their movement patterns or their clothes in between recordings. Machado \etal \cite{Machado2013} record several objects and models with the Kinect camera and let them annotate by human observers. The meshes are presented in various formats with the task to identify the object from the recorded shape. Luber \etal \cite{luber2011people} present a pedestrian dataset captured with three Kinects which are placed such that their viewing cones do not interfere. The dataset is annotated in that the position of each pedestrian is bounded by a rectangle in the input views. Their dataset contains of walking and standing pedestrians seen from different orientations and with different levels of occlusions. Silberman \etal \cite{Silberman:ECCV12} present a dataset consisting of 1449 labeled pairs of aligned RGB and depth images captured in indoor environments, such as bathrooms, basements, bedrooms, kitchens and playrooms. It includes the accelerometric data for each frame and also features a toolbox implemented in matlab that includes useful functions for manipulating the data and labels. Anand \etal \cite{anand2013contextually} captured several indoor environments and labeled the depth data. They also present in bag files the output of RGBDSLAM for each scene, e.g. for each timestamp a transform-matrix for that frame that transforms the camera from the first frame accordingly. Janoch \etal \cite{janoch2013category} show a large dataset annotated with the help of Amazon's Mechanical Turk consisting of indoor environment items like chairs, monitors, cups, bottles, bowls, keyboards, mouses or phones. They do not provide additional accelerometer data. Dataset consistiong of faces of 52 people (14 females, 38 males) captured with the Microsoft Kinect has been presented by Huynh \etal \cite{huynh2013efficient}. The faces are captured
in nine different conditions (neutral face, smile, mouth open, face in left profile, face in right profile, partial occlusion of face parts, changing lighting conditions). They do not include the accelerometric data. Defined landmark points were manually identified in the input images. In their work about motion recognition Sung \etal  \cite{sung2011human} provide depthmaps and skeletons for four subjects (two male, two female, one left-handed) who were asked to perform different high-level activities, like making cereal, arranging objects or having a meal. The activities are label and subclassified for movements like reaching, opening, placing, or scrubbing.

\section{Shortcomings}
\label{shortcomings}
The authors believe that, although there is already quite a remarkable amount of publicly available datasets based on capturings conducted with the Kinect, certain aspects in use of the sensor seem to be underrepresented. While already one paper is published~\cite{martinezkinect} that aims to extend the depth reconstruction capabilities from IR input stream data, a coherent dataset containing the IR data and additional ground truth depth information, e.g. from scene calibration or stereo, is missing. Also, arbitrary mesh reconstruction is in the datasets currently considered as byproduct of SLAM algorithms, Sect.~\ref{slam}, such that estimates with the accuracy of a few millimeters to a centimeter seem sufficient. However, recently publications have emerged to employ one or many Kinects for the accurate reconstruction of objects, e.g. based on depth, a combination of depth and texture cues in the RGB stream~\cite{miao2012texture} or from IR input stream~\cite{ou2011infrared}. The reconstructed objects in these setups need explicitly not necessary be purely opaque~\cite{lysenkov2012recognition,berger2011capturing}. A ground truth dataset with a high-resolution laser scan alongside input frames from Kinect (depth, RGB and IR) with a pose reconstruction of the sensor position would be highly desirable.

\section{Conclusion}
In this state-of the art report we have provided an overview over the publicly available datasets generated for benchmark with the Microsoft Kinect.
Several approaches, ranging from a steady single Kinect capturing setup over a moving Kinect in the scene to capturing setups that include multiple Kinects, have been discussed. The applicational context varied between SLAM, motion capturing and recognition. We have also phrased a critical view onto the diversity of current datasets with suggestions for extending the state of the art in benchmarks. With the deployment of the new \emph{Kinect One} in the near future the authors assume that in the next years the amount of publicly available benchmark datasets will increase significantly. 

{\small
\bibliographystyle{ieee}
\bibliography{egbib}
}
\begin{table*}
\centering
\begin{tabular}{|p{3cm}|p{3cm}|p{1.5cm}|p{1.5cm}|p{3cm}|p{3cm}|}
\hline
\textbf{Author}&\textbf{Intended Application}& \textbf{Datasize} & \textbf{Acce\-le\-ro\-me\-ter Data} & \textbf{Annotated}&\textbf{Link}\\
\hline
Glocker \etal \cite{Glocker2013}&SLAM& 151MB& No& Camera Path generated with KinectFusion \cite{izadi2011kinectfusion}&http://research.micro
soft.com/en-us
/projects/7-scenes/\\
\hline
Lieberknecht \etal \cite{lieberknechtIsmar11} &SLAM& $\approx$ 100 kB& No& No&https://www.dropbox
.com/sh/
1kyhns6s1xpbmzw/
RQKaYqdp7B/videos\\
\hline
Sturm \etal \cite{sturm2011towards} &SLAM& 50GB& Yes& Ground truth pose via external markers tracked with motion capturing system&https://cvpr.in.tum.de/ research/
datasets/rgbd-dataset\\
\hline
Anand \etal \cite{anand2013contextually} &Object Recognition& $\approx$ 7.6GB& Yes& Annotated Depth images&http://pr.cs.cornell.edu
/sceneunderstanding
/data/data.php
\\
\hline
Barbosa \etal \cite{barbosa2012re} &Object Recognition& 456 MB& No& Skeleton and Meshes&http://www.iit.it/en/
datasets/rgbdid.html
\\

\hline
Huynh \etal \cite{huynh2013efficient} &Object Recognition& no information & No& Faces labeled in input data&
http://rgb-d
.eurecom.fr/\\

\hline
Janoch \etal \cite{janoch2013category} &Object Recognition& 793 MB& No& Objects labeled in input data&
http://www.eecs.
berkeley.edu/\~
~allie
/VOCB3DO.zip\\
\hline
Lai \etal \cite{lai2011large} &Object Recognition& 84GB& No& Objects labeled in input data&http://www.cs.
washington.edu/
rgbd-dataset\\
\hline
Liu \etal \cite{liu2013learning} &Object Recognition& $\approx $1GB& No& Hand gestures labeled in input data&
http://lshao.staff.shef.
ac.uk/data/
SheffieldKinect
Gesture.htm\\
\hline
Luber \etal \cite{luber2011people} &Object Recognition& 2 GB& No& Pedestrians labeled in input data&
http://www.informatik
.uni-freiburg.de
/\~ ~spinello/sw
/rgbd\_people
\_unihall.tar.gz\\
\hline
Machado \etal \cite{Machado2013} &Object Recognition& 24.5MB& No& Objects Labeled&
https://dl.dropbox.com/
u/4151663/OR/
Dataset/test\%20set.zip\\
\hline
Negin \etal \cite{negin2013decision} &Object Recognition& 142GB& No& Motion Files containing the tracked joints&
http://vpa.sabanciuniv
.edu/databases/
WorkoutSU-10/
MinimalDataset.rar\\
\hline
Silberman \etal \cite{Silberman:ECCV12} &Object Recognition& 428GB& Yes& Labeled Depth Dataset&
http://cs.nyu.edu/
\~ ~silberman/datasets/
nyu\_depth\_v2.html\\
\hline
Sung \etal \cite{sung2011human} &Object Recognition& $\approx $13.8GB& No& Skeleton and activity/reachability lables&http://pr.cs.cornell
.edu/humanactivities
/data.php
\\
\hline
\end{tabular}
\caption{\label{overviewtable} Overview table for the benchmark datasets that are publicly available. We compared properties like data size (third row), the availability of the accelerometric data (fourth row) and the amount of annotion for ground truth (fifth row). For all datasets we listed the link under which they are publicly available. However, some datasets may require the request for login data.}
\end{table*}
\end{document}